\documentclass[10pt,twocolumn,letterpaper]{article}
\usepackage{subfigure}
\usepackage{bm}

\usepackage{algorithm}
\usepackage{algorithmic}

\usepackage{tabularx}
\usepackage{multirow}
\newcolumntype{C}[1]{>{\centering\arraybackslash\vspace{0pt}}m{#1}}
\newcolumntype{Y}{>{\centering\arraybackslash}X}

\usepackage{iccv}

\definecolor{iccvblue}{rgb}{0.21,0.49,0.74}
\usepackage[pagebackref,breaklinks,colorlinks,allcolors=iccvblue]{hyperref}

\def\paperID{*****} 
\def\confName{ICCV}
\def\confYear{2025}

\title{Task-Specific Generative Dataset Distillation with Difficulty-Guided Sampling}

\author{Mingzhuo Li$^1$ \quad Guang Li$^1$\thanks{Correspondence to: Guang Li (guang@lmd.ist.hokudai.ac.jp)} \quad Jiafeng Mao$^2$ \quad Linfeng Ye$^3$ \quad Takahiro Ogawa$^1$ \quad Miki Haseyama$^1$ \\ \\
$^1$Hokkaido Univerisy \quad  $^2$The University of Tokyo \quad $^3$University of Toronto\\
}

\begin{document}

\maketitle
\begin{abstract}
To alleviate the reliance of deep neural networks on large-scale datasets, dataset distillation aims to generate compact, high-quality synthetic datasets that can achieve comparable performance to the original dataset. The integration of generative models has significantly advanced this field. However, existing approaches primarily focus on aligning the distilled dataset with the original one, often overlooking task-specific information that can be critical for optimal downstream performance. In this paper, focusing on the downstream task of classification, we propose a task-specific sampling strategy for generative dataset distillation that incorporates the concept of difficulty to consider the requirements of the target task better. The final dataset is sampled from a larger image pool with a sampling distribution obtained by matching the difficulty distribution of the original dataset. A logarithmic transformation is applied as a pre-processing step to correct for distributional bias. The results of extensive experiments demonstrate the effectiveness of our method and suggest its potential for enhancing performance on other downstream tasks. The code is available at \url{https://github.com/SumomoTaku/DiffGuideSamp}.
\end{abstract}

\section{Introduction}
\label{sec:intro}
With the rapid advancement of deep learning, deep neural networks have gained significant attention due to their extensive applications across various domains, particularly in computer vision \cite{Gaurav2023DLsurvey}. However, these networks typically rely on large-scale datasets to obtain high performance, which results in extended training times that often span several hours or even days, and substantial demands on computational resources \cite{Mustafa2023DLsurvey}. Moreover, the storage and management of massive datasets involve considerable time and financial costs. Dataset distillation \cite{wang2018datasetdistillation} has emerged as a promising solution to mitigate these challenges by distilling the original dataset into a compact and high-quality synthetic dataset, which can train models to achieve performance comparable to that obtained using the original dataset.

\begin{figure*}[t]
        \centering
        \includegraphics[width=17cm]{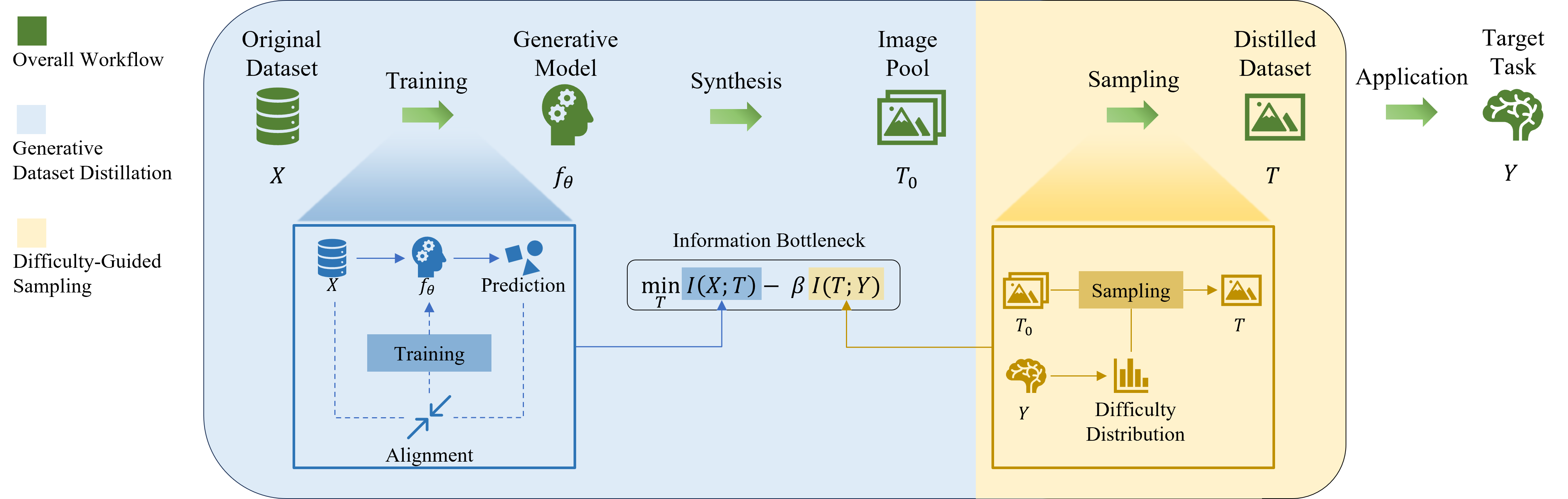}
        \caption{The workflow of the proposed method with the overall linear process marked as green. The generation of the image pool is indicated in blue, with optimization strategies aligning the original and distilled distributions. The sampling of the distilled dataset is indicated in orange, using the difficulty distribution that originates from the target task. The two components focus on different aspects of the Information Bottleneck optimization objective and are expected to function complementarily to enhance overall performance.}
        \label{workflow}
\end{figure*}

Since its introduction, dataset distillation has attracted significant attention, with a growing number of studies contributing to its rapid advancement \cite{li2022awesome, liu2025survey}. Current dataset distillation methods can be broadly categorized into non-generative and generative ones. Traditional non-generative methods aim to optimize a fixed set of synthetic images, with the size determined by image-per-class (IPC). The optimization is achieved by aligning specific training targets with those derived from the original dataset, under the assumption that models with similar alignment behavior will achieve comparable performance on downstream tasks. Various alignment targets have given rise to different methods, including gradient/trajectory matching \cite{zhao2021graMatch, cazenavette2022traMatch, li2023ddpp, li2024iadd},  distribution/feature matching \cite{zhao2023DM, sajedi2023disMatch, li2025hyperbolic, cui2025disMatch}, and kernel-based methods \cite{nguyen2021kerBase, chen2024kerBase}.

\par
In contrast, generative dataset distillation methods utilize generative models \cite{zhao2022genDD, cazenavette2023genDD, li2024generative, li2025generative} to produce high-quality synthetic images, which is made feasible by embedding knowledge of the dataset into the model. This modification offers the flexibility to generate datasets of any size on demand, effectively removing the constraint of IPC and reducing time costs, which is particularly beneficial for scenarios such as continual learning \cite{masarczyk2020reducing, gu2024continueLearning}, federated learning \cite{li2023sharing, jia2024fedratedLearning}, privacy preservation \cite{li2020soft, li2022compressed, zheng2024privacyPreservation}, and neural architecture search \cite{ding2024architectureSearch}. Among these models, diffusion models, such as Imagen \cite{saharia2023imagen} and Stable Diffusion \cite{rombach2022StableDiffusion}, have shown exceptional promise for their robustness and adaptability, promoting increasing interest in leveraging them for effective dataset distillation \cite{gu2024minimax, su2024difDD, su2024diffusion}.

\par

While current generative dataset distillation methods have demonstrated promising performance, the approaches primarily focus on guiding the model with knowledge extracted from the original dataset, overlooking the information specific to the downstream task \cite{gu2024minimax, su2024difDD, li2025diversity}. This discrepancy between the training objective and the target task may lead to incomplete information during training, limiting the model's optimal performance. To address this issue, we propose leveraging the relevance of the distilled dataset concerning the downstream task, aiming to generate datasets with superior performance on the target task.

\par
In this paper, we focus on the downstream task of classification and introduce a difficulty-guided sampling to enhance the performance of generative dataset distillation. An image pool consisting of generated images is first obtained using a generative dataset distillation method with the optimization objective of aligning the diversity and representativeness between the original and distilled datasets. The final distilled dataset is selected by aligning the difficulty distribution of the image pool with that of the original dataset. As previous generative models tend to produce samples biased toward lower difficulty (i.e., easier samples), a pre-processing step of logarithmic transformation is introduced for distributional correction. Extensive experiments on various downstream models and datasets demonstrate the effectiveness of our proposed method. The contributions of this paper can be summarized as follows:
\begin{itemize}
    \item We propose a difficulty-guided sampling to utilize extra information related to the classification task,  achieving task-specific dataset distillation.

    \item We conduct sampling on an image pool following the difficulty distribution of the original dataset, and propose a logarithmic transformation to eliminate the bias of the image pool towards easy samples.
\end{itemize}

\section{Dataset Distillation with Difficulty-Guided Sampling}
This section is organized as follows. We begin by reviewing a widely adopted generative dataset distillation pipeline, which is based on aligning the distribution between the distilled and original datasets. We then present the detailed implementation of difficulty-guided sampling, supported by theoretical analysis. Finally, we illustrate the logarithmic transformation, which is designed to obtain effective sample selection. The workflow of the proposed method is shown in Fig. \ref{workflow}.

\subsection{Preliminary}
\label{2-1}

Latent diffusion models \cite{rombach2022LDM} operate in the latent space rather than directly in the pixel space, showing enhanced ability on abstract features. Given an image $\bm{x}$ from the original dataset $D$, it is first encoded into a latent vector $\bm{z}_0$ by the VAE encoder. A noisy latent $\bm{z}_t$ is then obtained by sequentially adding Gaussian noise $\epsilon \in \mathcal{N}(\bm{0}, \bm{I})$ to $\bm{z}_0$ over $t$ times as follows:
\begin{equation}
    \bm{z_t} = \sqrt{\overline{\alpha}_t} \bm{z}_0 + \sqrt{1 - \overline{\alpha}_t} \epsilon, 
\end{equation}
where $\overline{\alpha}_t$ denotes a hyper-parameter known as variance schedule. The diffusion model parameterized by $\theta$ is trained to predict the added noise $\epsilon$, conditioned on class information $\bm{c}$, which is obtained via a class encoder. The training objective minimizes the discrepancy between the predicted noise $\epsilon_{\theta}(\bm{z_t}, t, \bm{c})$ and the ground truth $\epsilon$ as follows:
\begin{equation}
    \mathcal{L}_{\text{diffusion}} = \arg \max_{\theta} {||\epsilon_{\theta}(\bm{z_t}, t, \bm{c}) - \epsilon||}_2^2.
\end{equation}
Once trained, the model is capable of generating images by iteratively denoising random noise, thereby achieving high-quality image synthesis.

\par

To leverage diffusion models for dataset distillation, Minimax \cite{gu2024minimax} introduces an approach that aims to maximize both the representativeness and diversity of the distilled dataset. Two auxiliary memory sets are constructed to facilitate the calculation, with representativeness memory $\mathcal{M}_{r}$ containing real images and diversity memory $\mathcal{M}_{d}$ containing generated images. Representativeness is defined as the similarity between the generated and original dataset, leading to the optimization objective as follows:
\begin{equation}
    \mathcal{L}_\text{repre} = \arg \max_{\theta} \min_{\bm{z}_{r} \in [\mathcal{M}_{r}]} \sigma(\bm{\hat{z}}_{\theta}(\bm{z}_t, \bm{c}), \bm{z}_{r}),
\end{equation}
where $\sigma(\text{·} \ , \ \text{·})$ denotes the cosine similarity and $\bm{\hat{z}}_{\theta}(\bm{z}_t, \bm{c})$ is the latent predicted by the diffusion model $f_{\theta}$ with input latent $\bm{z}_t$ conditioned by class vector $\bm{c}$.
Similarly, diversity is defined based on the dissimilarity among the generated images, with the optimization objective as follows:
\begin{equation}
    \mathcal{L}_\text{div} = \arg \min_{\theta} \max_{\bm{\hat{z}}_{g} \in [\mathcal{M}_{d}]} \sigma(\bm{\hat{z}}_{\theta}(\bm{z}_t, \bm{c}), \bm{\hat{z}}_{g}).
\end{equation}
By combining $\mathcal{L}_\text{div}$ and $\mathcal{L}_\text{repre}$ with the diffusion loss $\mathcal{L}_{\text{diffusion}}$, the model is guided to produce distilled datasets of higher quality, thereby improving the performance on downstream tasks.

\par

Although this method effectively leverages features from the original dataset, it overlooks information specific to the downstream task. This omission can lead to a mismatch between the optimization objective during training and the target downstream tasks, such as classification, limiting the optimal performance.

\subsection{Difficulty-Guided Sampling}
From the perspective of the Information Bottleneck (IB) principle, the objective of dataset distillation can be redefined as follows. For the original dataset $X$ and target downstream task $Y$, the goal is to find a compressed dataset $T$ that discards irrelevant details from $X$ while retaining the information relevant to $Y$. Since the original dataset is no longer used during the downstream application, $Y$ is conditionally independent of
$X$ given $T$, resulting in the Markov chain structure $X \to T \to Y$. This satisfies the Markov assumption required by IB, leading to the objective as follows:
\begin{equation}
    \mathcal{L}_\text{IB} = \min_{T} I(X;T) - \beta \ I(T;Y),
\end{equation}
where $I(X;T)$ and $I(T;Y)$ denote the mutual information between $X$ and $T$, and between $T$ and $Y$, respectively. And $\beta$ is a Lagrange multiplier. The former part improves the level of compression, while the latter part enhances the predictability of the target. Balancing these two objectives helps construct distilled datasets that are both compact and effective.

\par

Recent generative dataset distillation methods primarily focus on optimizing the distribution of the distilled dataset concerning the original dataset. For example, the aforementioned Minimax enhances diversity and representativeness, while MGD3 \cite{chan-santiago2025mgd3} guides the denoising process toward desired distributional regions. These approaches can be broadly categorized as efforts to extract more features from the original dataset, making the distilled dataset resemble the original distribution. Since the original dataset inherently contains rich information, including labels for classification, such efforts implicitly benefit various downstream tasks. In other words, these approaches implicitly improve $I(T; Y)$ by explicitly improving $I(X; T)$, and the overall performance comes from the balancing of the two factors.
In the absence of task-specific considerations, the enhancement of $I(T; Y)$ is limited to the original dataset's inherent information, which may result in potentially suboptimal performance on the specific downstream task.

\par

To address this issue, we propose incorporating task-specific information to leverage the relevance between the distilled dataset $T$ and the target task $Y$, explicitly improving $I(T; Y)$ for better performance. Inspired by the findings of Wang et al. \cite{wang2024difficulty}, which demonstrate the effectiveness of controlling sample difficulty for dataset enhancement, we introduce difficulty as a proxy to quantify the information content for classification task.

\par

The difficulty of an image $\mathcal{D}_x$ is defined as the inverse of the confidence $P$ assigned to the correct class $y_{true}$ predicted by a pre-trained classification model $f_{\theta}$ as follows:
\begin{equation}
    \mathcal{D}_x = 1 - P_{f_{\theta}}(y_{true} | x).
\end{equation}
As illustrated in Fig.~\ref{workflow}, an image pool with a total size of $n \times \text{IPC}$is first constructed by collecting distilled images generated by the distillation pipeline of Minimax. The difficulty of each image in the image pool is then computed to serve as additional task-specific information. The sampling is then performed over the image pool following a specific sampling distribution.

\par

Assuming the original dataset represents the ground truth for optimal performance, we hypothesize that a distilled dataset exhibiting a similar difficulty distribution to the original one may yield improved performance. Consequently, the sampling distribution is obtained by scaling the difficulty distribution of the original dataset to match the IPC. The effectiveness of this scaling-based sampling is supported by our experiments in Section~\ref{samp_distr}, where we compare its performance with several pre-defined sampling distributions.

\begin{figure*}[t]
        \centering
        \includegraphics[width=17cm]{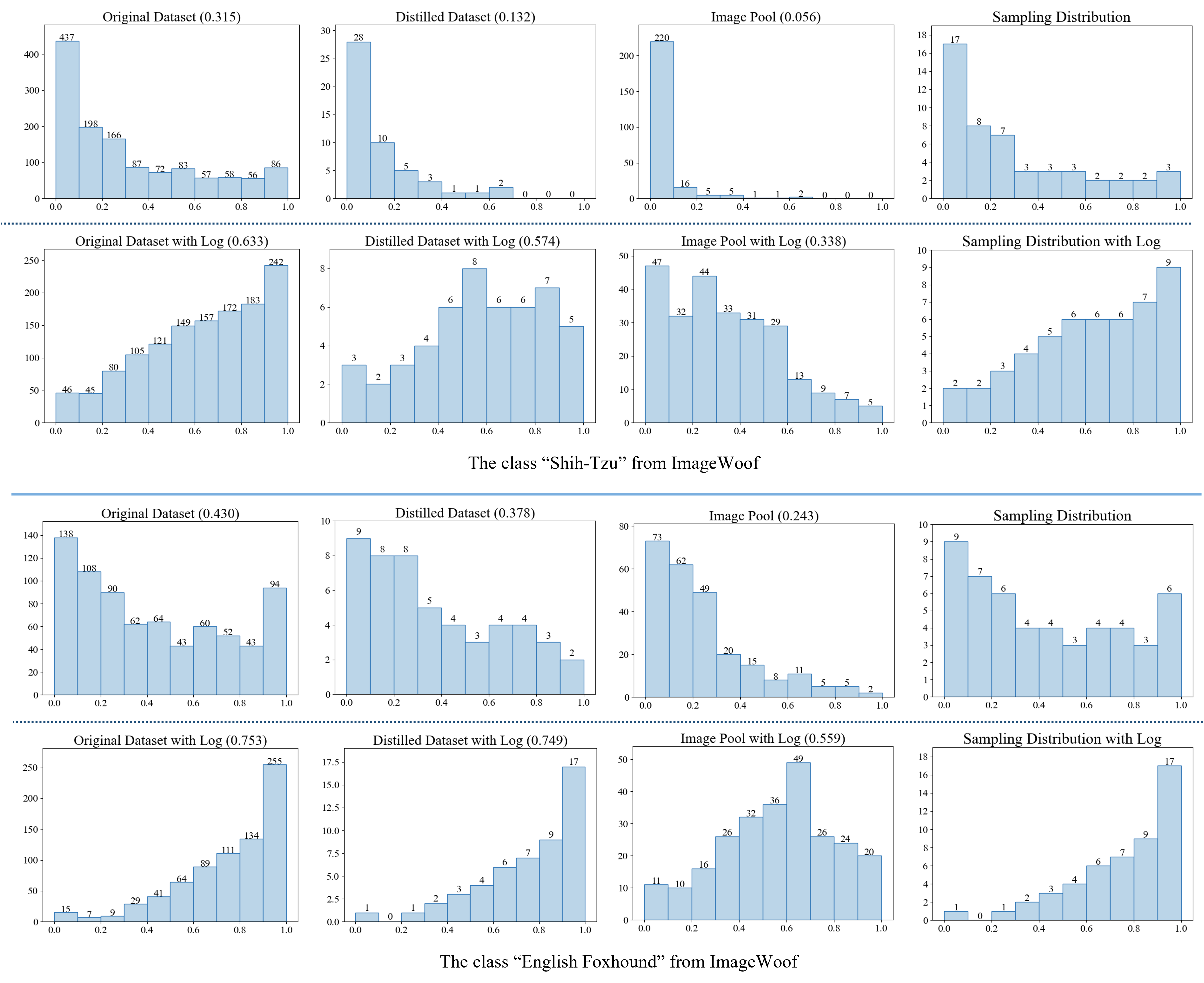}
        \caption{The difficulty distributions of different datasets, in the example of two classes of ImageWoof. The x-axis represents difficulty intervals, while the y-axis indicates the number of images per interval. The average difficulty of the dataset is annotated in the title. The lower and upper row shows the sampling process with and without logarithmic transformation, respectively.}
        \label{fig_distr}
\end{figure*}

\subsection{Pre-processing of Logarithmic Function}
However, a notable bias exists between the difficulty distributions of the original and generated datasets \cite{wang2024difficulty}. Distilled datasets, particularly those obtained by generative models, are also subject to the bias, tending to contain a higher proportion of easy samples, as illustrated in Fig.~\ref{fig_distr}. This imbalance hinders the coverage of the sampling distribution in certain areas, distorting the difficulty distribution of the distilled dataset, necessitating additional corrective steps.

\par
To address this issue, a logarithmic transformation is applied to facilitate the alignment of the difficulty distributions of both the original dataset and the image pool to enable better sampling. The target of transformation is selected as the uniform distribution following the idea that classification models benefit from balanced data. Due to the observation that many images tend to cluster around similar difficulty values, particularly in the lower and upper extremes, directly applying the logarithmic function may amplify the influence of extreme values, affecting the overall stability.

\par

Hence, thresholding at both the start and end of the original difficulty distribution $P_X(n)$ is introduced to stabilize the transformation and prevent the dominance of extreme values. The clipped distribution $P'_X(n)$ is obtained as follows:
\begin{equation}
    P'_X(n)= H(n - b) \ P_X(n) \ H(N - n - t) + \epsilon, 
\end{equation}
where $b$ and $t$ denote the bottom and top thresholds, respectively. $N$ is the size of $P_X(n)$,  $H(n)$ is the Heaviside step function and $\epsilon$ is a small value to avoid mathematic error. To keep the range between $0$ and $1$, the logarithmic transformation $f$ is defined as follows:
\begin{equation}
    f(P_X, b, t)= \frac{\ln (P_X'(n) / \min(P_X'(n)))}{\ln (\max(P_X'(n)) / \min(P_X'(n)))}. 
\end{equation}

\begin{table*}
    \centering
    \footnotesize
    \renewcommand{\arraystretch}{1.5}
    
    \tabcolsep=5pt
    \caption{Comparison of downstream validation accuracy with other SOTA methods on ImageWoof. The results are obtained with ResNetAP-10. The best results are marked in bold.}
    \label{exp_main}
    \begin{tabular}{C{40pt}| C{45pt}| C{30pt} C{30pt} C{30pt}| C{30pt} C{30pt} C{30pt} C{30pt} C{40pt}| C{40pt}}
        \hline
        IPC (Ratio) & Test Model & Random & K-Center \cite{sener2017kcenter} & Herding \cite{Welling2009Herding} & DiT \cite{Peebbles2023DiT} & DM \cite{zhao2023DM} & IDC-1 \cite{kim2022IDC} & Minimax \cite{gu2024minimax} & Ours & Full Dataset \\ 
        \hline
         
        & ConvNet-6 & $24.3_{\pm 1.1}$ & $19.4_{\pm 0.9}$ & $26.7_{\pm 0.5}$ & $34.2_{\pm 1.1}$ & $26.9_{\pm 1.2}$ & $33.3_{\pm 1.1}$ & $34.1_{\pm 0.4}$ & \bm{$35.1_{\pm 0.5}$} & $86.4_{\pm 0.2}$ 
        \\ 
        10 (0.8\%) & ResNetAP-10 & $29.4_{\pm 0.8}$ & $22.1_{\pm 0.1}$ & $32.0_{\pm 0.3}$ & $34.7_{\pm 0.5}$ & $30.3_{\pm 1.2}$ & $37.3_{\pm 0.4}$ & $35.7_{\pm 0.3}$ & \bm{$37.4_{\pm 0.3}$} & $87.5_{\pm 0.5}$ 
        \\
        & ResNet-18 & $27.7_{\pm 0.9}$  & $21.1_{\pm 0.4}$ & $30.2_{\pm 1.2}$ & $34.7_{\pm 0.4}$ & $33.4_{\pm 0.7}$ & $36.9_{\pm 0.4}$ & $35.3_{\pm 0.4}$ & \bm{$35.9_{\pm 0.6}$} & $89.3_{\pm 1.2}$ 
        \\ 
        \hline
         
        & ConvNet-6 & $29.1_{\pm 0.7}$ & $21.5_{\pm 0.8}$ & $29.5_{\pm 0.3}$ & $36.1_{\pm 0.8}$ & $29.9_{\pm 1.0}$ & $35.5_{\pm 0.8}$ & $36.9_{\pm 1.2}$ & \bm{$38.1_{\pm 0.2}$} & $86.4_{\pm 0.2}$ 
        \\ 
        20 (1.6\%) & ResNetAP-10 & $32.7_{\pm 0.4}$ & $25.1_{\pm 0.7}$ & $34.9_{\pm 0.1}$ & $41.1_{\pm 0.8}$& $35.2_{\pm 0.6}$ & $42.0_{\pm 0.4}$ & $43.3_{\pm 0.3}$ & \bm{$45.5_{\pm 0.4}$} & $87.5_{\pm 0.5}$ 
        \\ 
        & ResNet-18 & $29.7_{\pm 0.5}$ & $23.6_{\pm 0.3}$ & $32.2_{\pm 0.6}$ & $40.5_{\pm 0.5}$ & $29.8_{\pm 1.7}$ & $38.6_{\pm 0.2}$ & $40.9_{\pm 0.6}$ & \bm{$43.4_{\pm 1.0}$} & $89.3_{\pm 1.2}$ 
        \\ 
        \hline
         
        & ConvNet-6 & $41.3_{\pm 0.6}$ & $36.5_{\pm 1.0}$ & $40.3_{\pm 0.7}$ & $46.5_{\pm 0.8}$ & $44.4_{\pm 1.0}$ & $43.9_{\pm 1.2}$ & $51.4_{\pm 0.4}$ & \bm{$52.0_{\pm 0.6}$} & $86.4_{\pm 0.2}$ 
        \\ 
        50 (3.8\%) & ResNetAP-10 & $47.2_{\pm 1.3}$ & $40.6_{\pm 0.4}$ & $49.1_{\pm 0.7}$ & $49.3_{\pm 0.2}$ & $47.1_{\pm 1.1}$ & $48.3_{\pm 1.0}$ & $54.4_{\pm 0.6}$ & \bm{$57.1_{\pm 0.9}$} & $87.5_{\pm 0.5}$ 
        \\ 
        & ResNet-18 & $47.9_{\pm 1.8}$ & $39.6_{\pm 1.0}$ & $48.3_{\pm 1.2}$ & $50.1_{\pm 0.5}$ & $46.2_{\pm 0.6}$ & $48.3_{\pm 0.8}$ & $53.9_{\pm 0.6}$ & \bm{$54.9_{\pm 0.1}$} & $89.3_{\pm 1.2}$ 
        \\
        \hline
    \end{tabular}
\end{table*}

\par

While the introduction of thresholds helps to produce a more balanced difficulty distribution, it also introduces distortion from artificially modifying some values. The Kullback-Leibler (KL) divergence is introduced to measure distribution-level differences, assisting in the determination of the appropriate clipping level. With the target of being similar to both the uniform distribution $\mathcal{U}$ and the original difficulty distribution $P_X(n)$, the optimal threshold value is pinpointed as follows:
\begin{align}
\begin{split}
    b^*, t^* = \arg \min_{b, t} (\lambda \ D_\text{KL}(f(P_X, b, t) || P_X) \\
     + (1 - \lambda) \ D_\text{KL}(f(P_X, b, t) || \mathcal{U})),
\end{split}
\end{align}
where $D_\text{KL}(P||Q)$ denotes the KL divergence of P from Q and $\lambda \in [0, 1]$ is a weighting factor controlling the trade-off between uniformity and similarity.

\par

Through the above procedure, we obtain a distilled dataset that matches the difficulty distribution of the original dataset. In the specific downstream task of classification, it incorporates additional task-relevant information and is therefore expected to have improved performance compared to existing approaches.

\section{Experiments}
\label{sec_exp}

\subsection{Datasets and Evaluation}
To validate the effectiveness of the proposed method, extensive experiments are conducted on three 10-class subsets from the full-sized ImageNet \cite{jia2009imageNet} dataset: ImageWoof \cite{fastai20imageNettte}, ImageNette \cite{fastai20imageNettte}, and ImageIDC \cite{kim2022IDC}. These subsets differ in the difficulty of classes for classification, with ImageWoof being the most challenging one, consisting of 10 specific dog breeds. ImageNette consists of 10 specific classes that are easy to classify, and ImageIDC contains 10 classes randomly selected from ImageNet. We evaluate the classification accuracy of the proposed method and compare with several SOTA methods, including dataset selection methods like Random, K-Center \cite{sener2017kcenter} and Herding \cite{Welling2009Herding}, non-generative dataset distillation methods like DM \cite{zhao2023DM} and IDC-1 \cite{kim2022IDC}, and generative dataset distillation methods like DiT \cite{Peebbles2023DiT} and Minimax \cite{gu2024minimax}. The models for validation include ConvNet-6 \cite{gidaris2018convNet}, ResNet-18 \cite{he2016resNetAP}, and ResNet-10 with average pooling (ResNetAP-10) \cite{he2016resNetAP}, with a learning rate of 0.01 and top-1 accuracy being reported.

\begin{table}
    \centering
    \footnotesize
    \renewcommand{\arraystretch}{1.5}
    \caption{Comparison of downstream validation accuracy with other SOTA methods on different ImageNet subsets. The results are obtained with ResNetAP-10. The best results are marked in bold.}
    \label{exp_dataset}
    \begin{tabular*}{\linewidth}{c|c|cccc}
        \hline
        & IPC & Random & DiT \cite{Peebbles2023DiT} & Minimax \cite{gu2024minimax} & Ours \\

        \hline
        \multirow{3}*{\rotatebox{90}{ImageNette}} & 10 & $54.2_{\pm 1.6}$  & $59.1_{\pm 0.7}$ & $59.8_{\pm 0.3}$ & \bm{$61.5_{\pm 0.9}$}
        \\
        & 20 & $63.5_{\pm 0.5}$ & $64.8_{\pm 1.2}$ & $66.3_{\pm 0.4}$ & \bm{$66.9_{\pm 0.5}$}
        \\
        & 50 & $76.1_{\pm 1.1}$ & $73.3_{\pm 0.9}$ & $75.2_{\pm 0.2}$ & \bm{$76.8_{\pm 0.7}$} 
        \\

        \hline
        \multirow{3}*{\rotatebox{90}{ImageIDC}} & 10  & $48.1_{\pm 0.8}$ & $54.1_{\pm 0.4}$ & $60.3_{\pm 1.0}$ & \bm{$61.6_{\pm 0.7}$}
        \\
        & 20 & $52.5_{\pm 0.9}$ & $58.9_{\pm 0.2}$ & $63.9_{\pm 0.4}$ & \bm{$64.3_{\pm 0.5}$}
        \\
        & 50 & $68.1_{\pm 0.7}$ & $64.3_{\pm 0.6}$ & $74.1_{\pm 0.2}$ & \bm{$74.2_{\pm 0.7}$}
        \\
        \hline
    \end{tabular*}
\end{table}

\par

The image pool is created using the Minimax pipeline with its default parameters and settings. For the diffusion model, a pre-trained DiT \cite{Peebbles2023DiT} with Difffit \cite{xie2023Difffit} for fine-tuning, and VAE \cite{kingma2013VAE} as the encoder. The input image is randomly arranged and transformed to $256 \times 256$ pixels. The number of denoising steps in the sampling process is 50. The distillation process lasts for 8 epochs with a mini-batch size of 8. An AdamW with a learning rate of 1e-3 is adopted as the optimizer. A ResNet-50 trained on the full ImageNet dataset is used as the pre-trained model for obtaining the difficulty scores. Each experiment is repeated 3 times, and the mean value and standard deviation are recorded. 

\subsection{Benchmark Results}
\begin{figure*}[t]
        \centering
        \includegraphics[width=17cm]{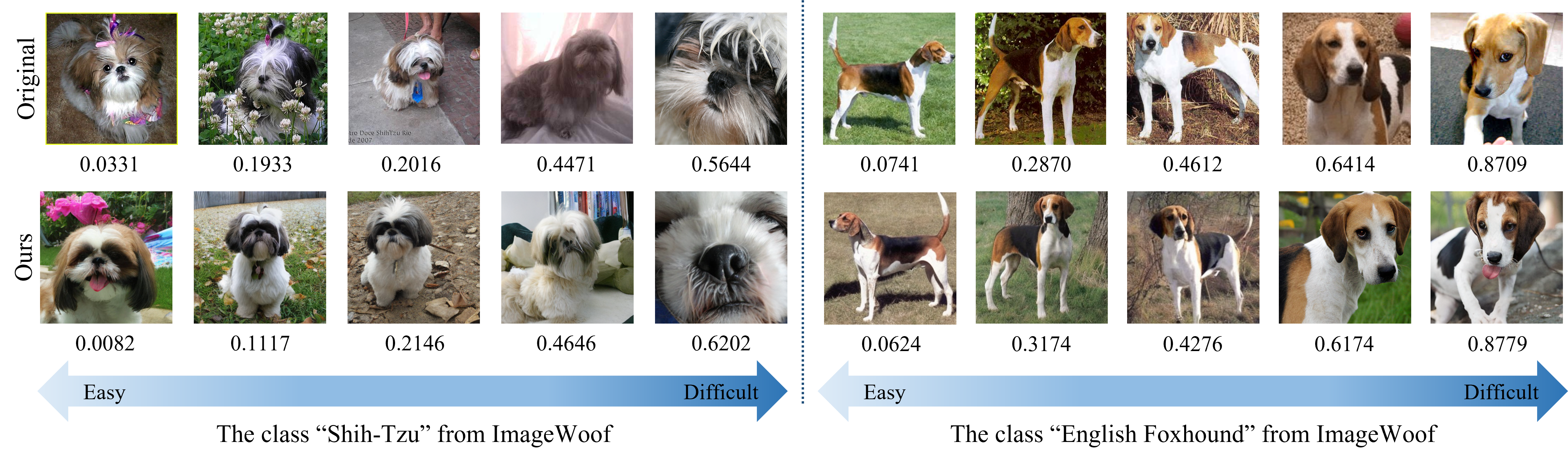}
        \caption{Visualization of images of the original and distilled dataset with difficulty scores.}
        \label{fig_visual}
\end{figure*}

Firstly, we compare the proposed method on the ImageWoof with different classification models and various IPC settings to show the method's cross-architecture effectiveness. As shown in Table~\ref{exp_main}, our method demonstrates superior accuracy across all experiments, especially in high IPC settings, proving the method's ability to enhance the task-specific performance of dataset distillation. 

\par

Then, we verify the generalization performance of the proposed method by conducting experiments on various datasets. As shown in Table~\ref{exp_dataset}, the performance trend observed on ImageNette and ImageIDC generally corresponds with those on ImageWoof, with the best performance demonstrated in most experiments. 

\par

To provide an intuitive understanding of our method, we illustrate the difficulty distributions of different datasets during the sampling process in Fig.~\ref{fig_distr}, using two example classes ``Shih-Tzu" (n02086240) and ``English Foxhound" (n02089973) from ImageWoof. As shown in the figure, the image pool obtained by generative models exhibits a strong bias toward easy samples, failing to reflect the difficulty characteristics of the original dataset. As a result, distilled datasets using the original distribution have many difficulty intervals remaining unrepresented, reducing the effects of sampling. By contrast, the logarithmic transformation flattens the difficulty distribution of the image pool, facilitating the sampling of images matching the target distribution. However, it also alters the original dataset’s difficulty distribution after transformation, highlighting the need for further discussion on the impact and potential solutions, such as adjusting transformation parameters.

\par

We also visualize the images of the aforementioned two classes in both the original and distilled datasets in Fig.~\ref{fig_visual}, along with their corresponding difficulty scores. The comparison reveals that the distilled dataset contains images of various difficulties and visual characteristics, indicating good sample diversity. Additionally, images of the same difficulty share some common features, suggesting the potential factors that contribute to the difficulty.

\subsection{Sampling Distribution}
\label{samp_distr}
\begin{figure}[t]
    \centering
    \includegraphics[width=\linewidth]{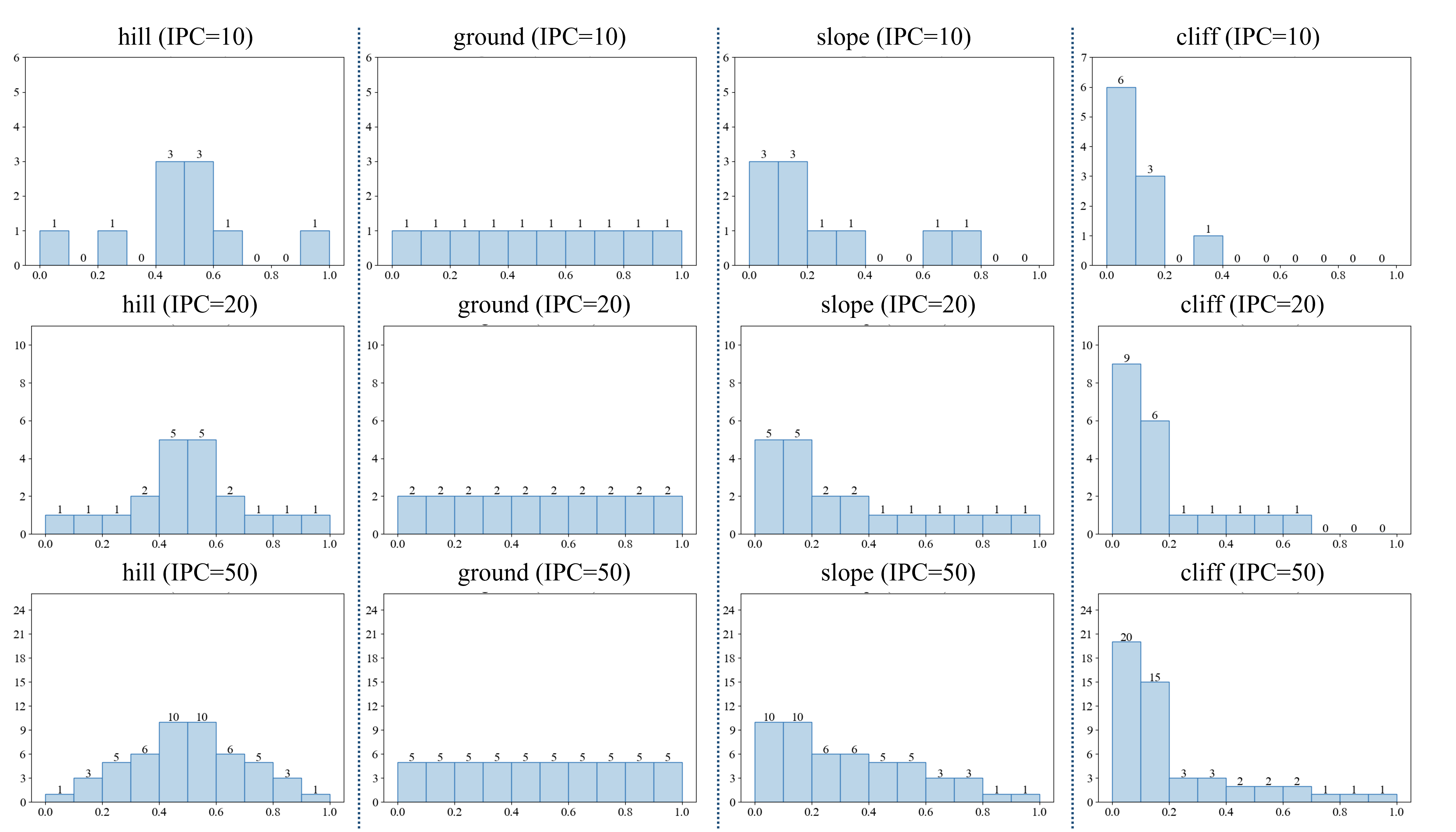}
    \caption{Visualization of pre-defined sampling distributions under different IPC settings, with increasing proportions of easy samples from left to right. The x-axis represents difficulty intervals, while the y-axis indicates the desired number of selected images per interval.
    }
    \label{fig_samp_distr}
\end{figure}

When obtaining the sampling distribution, we hypothesize that a distribution similar to the original dataset contributes to enhanced performance. We validate the hypothesis in Table~\ref{exp_samp_distr}, where we compare the downstream performance with various pre-defined sampling distributions, with ``scale" denoting the strategy of scaling the difficulty distribution of the original dataset. As illustrated in Fig.~\ref{fig_samp_distr}, the four pre-defined distributions ``hill", ``ground", ``slope" and ``cliff" are named according to their shapes, including increasing proportions of easy samples. 

\par

\begin{table}
    \centering
    \footnotesize
    \renewcommand{\arraystretch}{1.5}
    \caption{Comparison of downstream validation accuracy for different sampling distributions with the best results marked in bold. The label ``scale" refers to the strategy of scaling the difficulty distribution of the original dataset. The results are obtained with ResNetAP-10 on ImageWoof. The best results are marked in bold.}
    \label{exp_samp_distr}
    \centering
    \begin{tabularx}{\linewidth}{Y|YYY}
        \hline
        Distribution &  IPC = 10 & IPC = 20 & IPC = 50
        \\
        \hline
        Hill & $35.8_{\pm 0.2}$ & $41.7_{\pm 0.3}$ & $56.9_{\pm 0.5}$
        \\
        Ground & $36.7_{\pm 0.7}$ & $42.7_{\pm 0.8}$ & $55.0_{\pm 0.6}$
        \\
        Slope & \bm{$37.8_{\pm 0.4}$} & $42.7_{\pm 0.7}$ & $56.1_{\pm 0.6}$
        \\
        Cliff & $37.4_{\pm 0.6}$ & $44.3_{\pm 0.3}$ & $56.6_{\pm 0.8}$
        \\
        Scale & $37.4_{\pm 0.3}$ & \bm{$45.5_{\pm 0.4}$} & \bm{$57.1_{\pm 0.9}$}
        \\
        \hline
    \end{tabularx}
\end{table}

Experimental results show that distilled datasets sampled using the ``scale" distribution achieve the best performance, likely due to class-wise differences in difficulty distributions within the dataset. Further comparisons among pre-defined sampling distributions reveal that smaller sampled datasets with a higher proportion of easier samples, as well as larger sampled datasets with a higher proportion of more difficult samples, tend to yield better performance. This finding suggests that adjusting the proportion of easy and difficult samples according to the IPC setting may lead to improved performance.

\subsection{Size of Image Pool}
Since the final distilled dataset is sampled from the image pool, its size can influence the overall performance, necessitating efforts to determine the appropriate size. To this end, we construct image pools of varying sizes of $n \times \text{IPC}$ and conduct experiments to identify the optimal value of $n$ for practical implementation.

\par

As shown in Table~\ref{exp_ip_size}, the classification accuracy varies with the size of the image pool. Based mainly on results under higher IPC settings, the size of $5 \times \text{IPC}$ yields the relatively best performance, and is therefore adopted in subsequent experiments. This behavior can be attributed to a trade-off: while a larger image pool increases diversity, it also introduces redundancy, especially in the context of concentrated difficulty distributions shown in Fig.~\ref{fig_distr}. Moreover, the size affects the selection of threshold parameters in the logarithmic transformation, which are also applied to the original dataset, resulting in different sampling distributions. 

\begin{table}
    \centering
    \footnotesize
    \renewcommand{\arraystretch}{1.5}
    \caption{Comparison of downstream validation accuracy for different sizes of image pool. The results are obtained with ResNetAP-10 on ImageWoof. The best results are marked in bold.}
    \label{exp_ip_size}
    \centering
    \begin{tabularx}{\linewidth}{Y|YYY}
        \hline
        Size &  IPC=10 & IPC=20 & IPC=50
        \\
        \hline
        2 $\times$ IPC & $37.9_{\pm 0.4}$ & $44.3_{\pm 0.3}$ & $55.8_{\pm 0.6}$
        \\
        3 $\times$ IPC & $35.9_{\pm 0.8}$ & $41.5_{\pm 0.5}$ & $56.7_{\pm 0.4}$
        \\
        4 $\times$ IPC & \bm{$38.4_{\pm 0.6}$} & $43.7_{\pm 0.5}$ & $55.4_{\pm 0.9}$
        \\
        5 $\times$ IPC & $37.4_{\pm 0.3}$ & \bm{$45.5_{\pm 0.4}$} & \bm{$57.1_{\pm 0.9}$}
        \\
        6 $\times$ IPC & $34.5_{\pm 0.8}$ & $42.7_{\pm 1.1}$ & $54.9_{\pm 1.3}$
        \\
        \hline
    \end{tabularx}
\end{table}

\section{Conclusion}
In this paper, we have proposed a difficulty-based sampling method to improve task-specific performance for dataset distillation. Unlike previous methods that focus on the information between the distilled and original datasets for optimization, we evaluate the information relevant to the target downstream task by using difficulty distribution to facilitate sampling, offering complementary optimization based on the Information Bottleneck principle. 
The proposed method achieves state-of-the-art performance in the specific task of classification in most experiments, verifying the effectiveness of difficulty-based sampling. Moreover, it also supports the effectiveness of task-specific information, suggesting its potential for enhancing performance on other downstream tasks. 

{
    \small
    \bibliographystyle{ieeenat_fullname}
    \bibliography{main}
}

\end{document}